\documentclass[10pt,twocolumn,a4paper]{esaAI}
\usepackage{float}
\title{Generative Design of Periodic Orbits in the Restricted Three-Body Problem }

\def\authorEmail{\\alvaro.francisco.gil@alumnos.upm.es}

\author[1]{Alvaro Francisco Gil\thanks{Corresponding author. E-Mail: \authorEmail}}
\author[2]{Walther Litteri}
\author[1]{Victor Rodriguez-Fernandez}
\author[1]{David Camacho}
\author[2]{Massimiliano Vasile}

\affil[1]{School of Computer Systems Engineering, Universidad Politécnica de Madrid, Spain}
\affil[2]{Aerospace Centre of Excellence, University of Strahclyde, Glasgow, United Kindom}

\begin{document}

\makeCustomtitle

\begin{abstract}
The Three-Body Problem has fascinated scientists for centuries and it has been crucial in the design of modern space missions. Recent developments in Generative Artificial Intelligence hold transformative promise for addressing this longstanding problem. This work investigates the use of Variational Autoencoder (VAE) and its internal representation to generate periodic orbits. We utilize a comprehensive dataset of periodic orbits in the Circular Restricted Three-Body Problem (CR3BP) to train deep-learning architectures that capture key orbital characteristics, and we set up physical evaluation metrics for the generated trajectories. Through this investigation, we seek to enhance the understanding of how Generative AI can improve space mission planning and astrodynamics research, leading to novel, data-driven approaches in the field.
\end{abstract}

\section{Introduction}

Recent advancements in Artificial Intelligence (AI) have significantly impacted the space sector \cite{furano2020towards, hassan2024application}. AI techniques have been applied in tasks such as space traffic management \cite{vasile2017, SANCHEZFERNANDEZMELLADO2021694}, space object characterization \cite{furfaro2019space,VASILE2023510}, satellite pose estimation \cite{kisantal2020satellite}, natural language processing for space mission design \cite{berquand2020space}, and spacecraft operations using large language models \cite{rodriguez2022language}. Research on large pre-trained models has also emerged in astronomy, with applications in both generative and discriminative tasks. Notable models include AstroCLIP \cite{lanusse2023astroclip}, which uses cross-modal contrastive learning for astronomical images and spectra, and ASTROMER \cite{donoso2022astromer}, a transformer-based model for creating representations of light curves in a self-supervised manner.

In astrodynamics, machine learning, especially deep learning, has seen significant growth. Models are used to learn guidance or control laws \cite{izzo2019learning}, and deep learning has been applied to design complex trajectories in multi-body dynamics \cite{bonasera2022deep}, the solution of two-point boundary value problems \cite{bin2022} and the classification of regular and chaotic motion \cite{celletti2022}. Self-supervised learning techniques have shown effectiveness in tasks such as conjunction screening and maneuver detection, revealing potential for data-driven approaches in orbit analysis \cite{stevenson2023selfsupervised}.

In the context of the Three-Body Problem (3BP), AI has been used to tackle complex challenges. Support vector machines classified trajectories in the circular restricted three-body problem \cite{li2015}, deep neural networks solved the chaotic three-body problem \cite{breen2020}, and artificial neural networks predicted periodic orbits in three-body systems with arbitrary masses \cite{liao2022}. However, generative AI techniques have only recently been applied to astrodynamics \cite{wilson2024,EP_X018288_1}. This paper presents some early results from the OrbitGPT project \cite{esa_activities_4000143521}.

\subsection{The Restricted Three-Body Problem}\label{sec:r3bp}

The Three-Body Problem (3BP), introduced by Newton in 1687 \cite{philosophiae}, involves predicting the motion of three celestial bodies interacting gravitationally. Euler and Lagrange made significant contributions in the 18th century, describing equilibrium points where a small, massless particle influenced by two larger bodies (primaries) can remain stationary. Euler identified three collinear points, while Lagrange added two more points, assuming the primaries move in circular motion about their center of mass \cite{lagrange}. These models are known as the Restricted Three-Body Problem (R3BP) and the Circular-Restricted Three-Body Problem (CR3BP).

The equations of motion (EOM) for the CR3BP are:

\begin{subequations}
 \label{eq:eom_CR3BP}
 \begin{align}
     \Ddot{x} &= 2\Dot{y} + U_x \\
     \Ddot{y} &= -2\Dot{x} + U_y \\
     \Ddot{z} &= U_z
 \end{align}
\end{subequations}

\begin{figure}[t]
    \centering
    \includegraphics[width=.79\columnwidth]{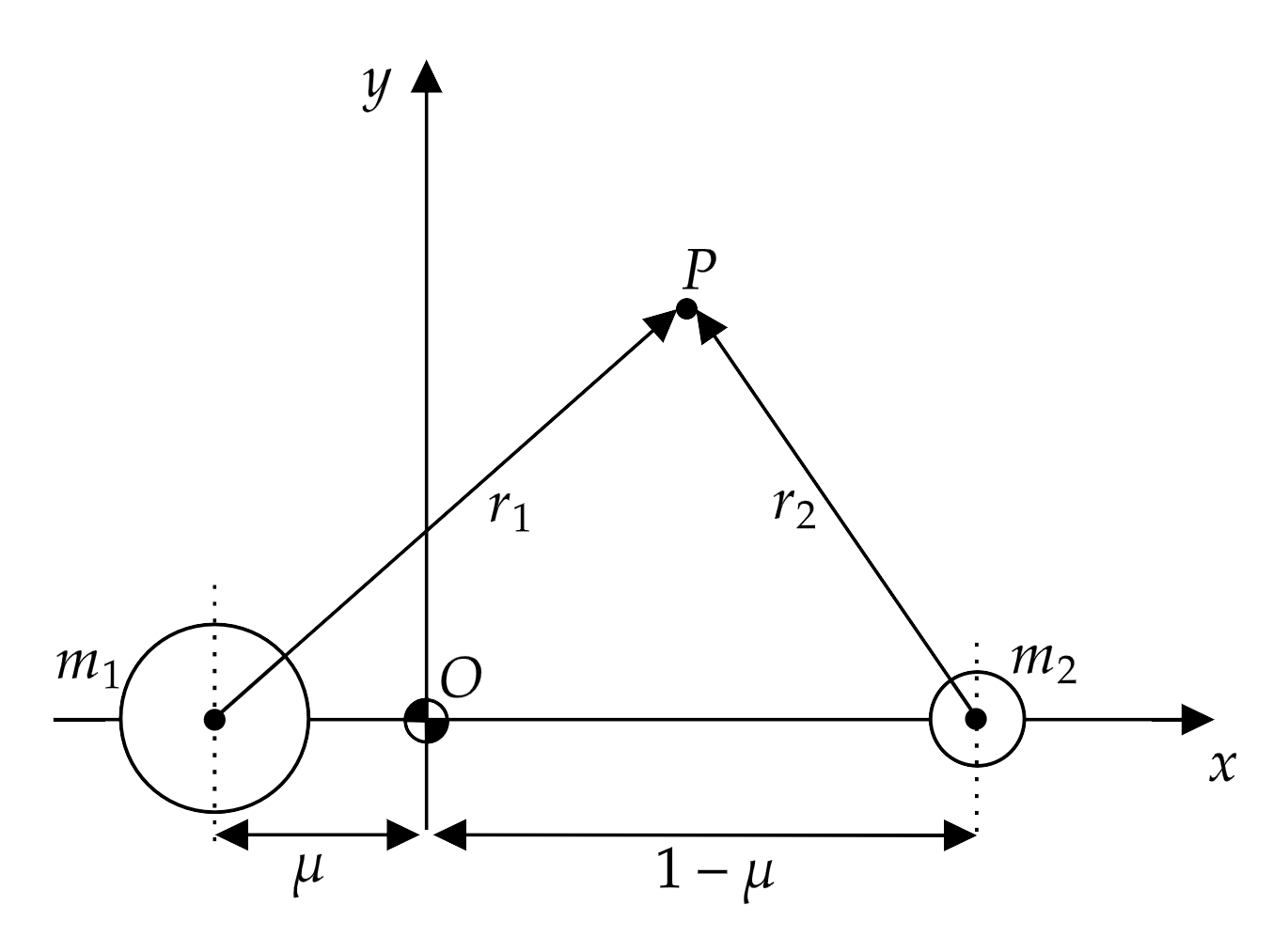}
    \caption{Geometry of the Circular-Restricted Three-Body Problem in a rotating reference frame.}
    \label{fig:cr3bp}
\end{figure}

The state vector \(\mathbf{X} = [x,y,z,\Dot{x},\Dot{y}, \Dot{z}]^T \in \mathbb{R}^6\) represents the position and velocity of the particle \(P\) relative to the center of mass in the non-inertial reference frame, as shown in Figure \ref{fig:cr3bp} \cite{threebodyprob}.

The state vector and its EOM are dimensionless, with the primaries at a fixed relative distance of 1 and rotating at an angular speed of 1. The potential function \(U\) is given by:

\begin{equation}
    U(x,y,z) = \frac{1}{2}(x^2 + y^2) + \frac{1-\mu}{r_1} + \frac{\mu}{r_2}
\end{equation}

where \(r_1\) and \(r_2\) are the distances between \(P\) and the two bodies. The components of its gradient, \(U_x, U_y, U_z\), are the forcing terms in Eqs. \ref{eq:eom_CR3BP}. The gravitational parameter \(\mu = m_2/(m_1 + m_2)\) defines the system's behavior, with \(0 \leq \mu \leq 0.5\). For the Earth-Moon system, \(\mu = 0.01215\).

\subsection{Periodic Orbits}

Let \(\mathbf{\Phi}(t,\mathbf{X}_0): \mathbb{R}\times\mathbb{R}^6 \rightarrow \mathbb{R}^6\) be the evolution function for the dynamical system described by the equations of motion (EOM) in Eqs. \ref{eq:eom_CR3BP}. This function determines the state vector \(\mathbf{X}(t)\) at any time \(t\) based on the initial state \(\mathbf{X}_0\): \(\mathbf{X}(t) = \mathbf{\Phi}(t,\mathbf{X}_0)\). A periodic orbit with period \(T\) is a trajectory where the state vector returns to its initial state after time \(T\): \(\mathbf{\Phi}(T,\mathbf{X}_0) = \mathbf{X}_0\).

Periodic and quasi-periodic orbits in the Circular Restricted Three-Body Problem (CR3BP) have fascinated scientists since the 19th century, especially with the advent of numerical computation. In 1892, Poincaré proved that infinite periodic solutions exist for the Three-Body Problem and highlighted the potential for chaotic motion \cite{poincare}. The discovery of unique families of periodic orbits in both two-dimensional and three-dimensional CR3BP began in the 1920s \cite{zhang}. Within any family of periodic trajectories, characteristics such as period, energy, and stability indicators vary continuously.

Interest in these orbits surged with the dawn of space missions, offering cost-effective and scientifically valuable opportunities. For instance, the International Sun/Earth Explorer 3 (ISEE-3) mission, launched in 1978, was the first to utilize a Halo orbit around the Lagrange point \(L_1\) in the Sun-Earth system \cite{isee3}. More recently, the James Webb Space Telescope, launched in December 2021, operates in a Halo orbit around the \(L_2\) point in the same system \cite{jwst}. NASA's CAPSTONE mission, launched in June 2022, uses a more stable Near Rectilinear Halo Orbit (NRHO) in the cislunar environment \cite{capstone}.

The analysis of periodic orbits in the CR3BP is a vibrant research area in aerospace science, contributing to our fundamental understanding of this dynamical system and enabling complex and efficient space missions.

\subsection{Generative AI}

Generative AI has become a transformative force across scientific and industrial domains, advancing data generation and pattern synthesis. Models like Generative Adversarial Networks (GANs) \cite{goodfellow2014generative}, Variational Autoencoders (VAEs) \cite{kingma2013auto}, and diffusion models \cite{ho2020denoising} have shown remarkable capabilities in producing realistic data, including images, speech, text, and time series. These models learn underlying data distributions to generate new, realistic instances.

Large pre-trained models, such as GPT-4 \cite{openai2024gpt4} and Llama2 \cite{touvron2023llama}, have revolutionized natural language processing (NLP) with their ability to understand and generate human-like text, enabling applications like conversational agents and automated content creation. In computer vision, models like DALL-E \cite{ramesh2021dalle} and Stable Diffusion \cite{rombach2022high} have demonstrated the power of generative AI in creating photorealistic images from textual descriptions, transforming our interaction with visual data.

\subsection{Variational Autoencoder}

A Variational Autoencoder (VAE) is a generative model that combines variational inference and neural networks to generate new data points. VAEs are useful for tasks like image synthesis, anomaly detection, and data compression, as introduced by Kingma and Welling \cite{kingma2013auto}.

The VAE consists of an encoder and a decoder, forming an autoencoder architecture with stochastic elements and probabilistic principles. The encoder transforms input data \(x\) into a latent representation \(z\), producing parameters (mean \(\mu(x)\) and standard deviation \(\sigma(x)\)) for the probability distribution \( q_{\phi}(z|x) \):
\[
q_{\phi}(z|x) = \mathcal{N}(z; \mu(x), \sigma(x))
\]
Here, \(\phi\) represents the encoder's parameters.

VAEs sample \(z\) from the distribution parameterized by \(\mu(x)\) and \(\sigma(x)\):
\[
z = \mu(x) + \sigma(x) \cdot \epsilon
\]
where \(\epsilon\) is a random variable from a standard normal distribution \(\mathcal{N}(0, 1)\).

The decoder reconstructs the input data \(x\) from \(z\), maximizing the likelihood of the data given the latent variable:
\[
p_{\theta}(x|z)
\]
Here, \(\theta\) represents the decoder's parameters.

The VAE is trained to maximize the Evidence Lower Bound (ELBO), which consists of the reconstruction loss and the KL-divergence:
\[
\mathcal{L}(\phi, \theta; x) = \mathbb{E}_{q_{\phi}(z|x)}[\log p_{\theta}(x|z)] - \text{KL}(q_{\phi}(z|x) \| p(z))
\]
where \(\text{KL}(q_{\phi}(z|x) \| p(z))\) is the Kullback-Leibler divergence between the encoder's output distribution and the prior distribution \(p(z)\).

\subsection{Research Motivation}

This paper is part of the OrbitGPT project \cite{esa_activities_4000143521}, which aims to apply generative AI to astrodynamics. The primary goal is to develop a large orbit model (LOM) that generates orbital trajectories with desired features, reducing the need for conventional design or orbit determination algorithms. This approach could revolutionize space mission design by generating new types of orbits, minimizing mission analysis costs, capturing past orbital knowledge, and training specialized models through synthetic data generation.

\section{Methodology}

\subsection{Dataset}

For this study, we use an extensive dataset precomputed by NASA, comprising 44,112 periodic initial conditions in the Earth-Moon CR3BP, classified into 40 families of orbits \cite{jpl}. The dataset includes various types of orbits, such as those located at the libration points (e.g., planar Lyapunov, Axial, Halo, and Vertical Orbits) and those developing around the entire system (e.g., Butterfly, Dragonfly, planar Distant Retrograde and Prograde Orbits, and Long Period Orbits).

The initial dataset from NASA consists only of the initial conditions of the orbits, along with their periods, stabilities, and Jacobi constants. To obtain the full vector of positions for each orbit, the initial conditions are integrated over time for one period using 100 nodes (N=100) with Matlab's ODE113 solver \cite{matlab_ode113}, with both absolute and relative tolerances set to \num{1e-13}.

The dataset is structured in a three-dimensional numpy array with a shape of data.shape = (num\_orbits, 7, num\_time\_points), where num\_orbits is 44,112, indexing each distinct orbit. The second dimension (7) contains seven scalar values for each orbit at every time point: position components (posX, posY, posZ), velocity components (velX, velY, velZ), and time. The third dimension, num\_time\_points, represents the number of time points at which data for each orbit is recorded, initially set to 100 nodes. The dataset is standardized using the min-max method for each scalar to facilitate efficient data handling and analysis.

\subsection{Model}

\begin{figure}[H]
    \centering
    \includegraphics[width=.99\columnwidth]{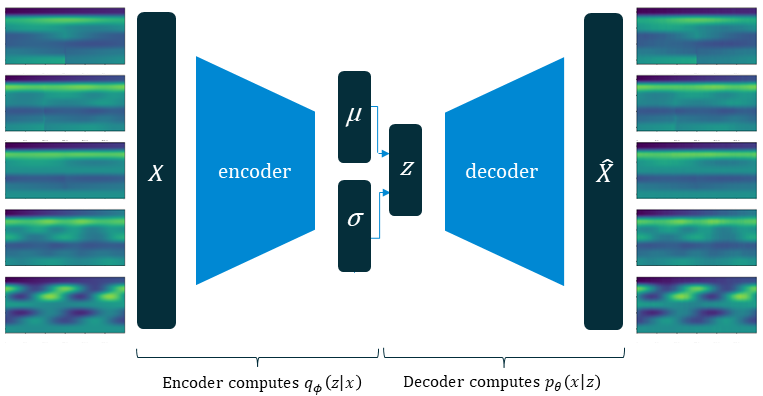}
    \caption[Schematic representation of an orbit Variational Autoencoder for orbit data]{Schematic representation of an orbit Variational Autoencoder architecture (VAE).
    }
    \label{fig_vae}
\end{figure}

The model trained on the orbit data is a Variational Autoencoder (VAE), as depicted in Figure~\ref{fig_vae}. This VAE utilizes a convolutional neural network (CNN) architecture specifically designed for time series data, implemented using the TSGM library \cite{nikitin2023tsgm}.

The encoder consists of five Conv1D layers, each followed by dropout layers to enhance the robustness of the model and prevent overfitting. After passing through the convolutional layers, the data is flattened and fed into two fully connected (dense) layers. The first dense layer has 512 neurons with ReLU activation, reducing the dimensionality of the feature space while retaining essential information. The second dense layer further reduces the feature space to 64 dimensions. This step ensures that the high-dimensional input data is compressed into a more manageable form before reaching the latent space. The output from these dense layers is then used to compute the mean (\( \boldsymbol{\mu} \)) and log-variance (\( \log(\boldsymbol{\sigma}^2) \)) of the latent variable distribution. These parameters define the latent space where the encoder maps the input orbit data.

The decoder follows an inverse architecture to the encoder. It first passes through dense layers with 64 and 512 neurons using ReLU activation, then a final dense layer reshapes the data for the five Conv1DTranspose layers. These layers, interspersed with dropout layers for robustness, reconstruct the input sequences, ending with a Conv1DTranspose layer with a sigmoid activation to produce the final output sequences matching the input dimensionality.

\subsection{Convergence}\label{sec:convergence}

The generative model described above may not produce fully physical solutions, but the generated trajectories can serve as initial guesses for trajectory optimization algorithms to generate actual physical periodic orbits.

A Multiple-Shooting (MS) algorithm iteratively adjusts a discretised trajectory $\{\mathbf{X}_i^0\}_{i=1}^{N}$ to satisfy a set of constraints:
\begin{equation}
    \mathbf{F}(\mathbf{\Bar{X}}) = \mathbf{0}
\end{equation}
Here, $\Bar{\mathbf{X}} \in \mathbb{R}^{N\times6 + (N-1)}$ includes $N$ states $\{\mathbf{X}_i^0\}_{i=1}^{N}$ and $N-1$ time intervals $dt_i = t_{i+1} - t_i, \  i=1,\ldots,N-1$. \\ The constraint vector $\mathbf{F} \in \mathbb{R}^{N\times6}$ enforces the respect of both the dynamical equations as in Eq. \ref{eq:dyn}, and the periodicity condition reported in Eq. \ref{eq:period}.
\begin{align}
\label{eq:dyn}
    \mathbf{F}_i &= \mathbf{X}_{i}^f - \mathbf{X}_{i+1}^0, \ i=1,\ldots,N-1 \\ \label{eq:period}
    \mathbf{F}_N &= \mathbf{X}_{1}^0 - \mathbf{X}_{N}^0
\end{align}
The notation $\mathbf{X}_i^f$ stands for the state vector obtained through the numerical integration of Eq. \ref{eq:eom_CR3BP} from the initial condition $\mathbf{X}_i^0$ over the time interval $dt_i$. \\
The algorithm then uses a Newton-Raphson method for iterative correction:
\begin{equation}
    \Bar{\mathbf{X}}^{j} = \Bar{\mathbf{X}}^{j-1} - \mathbf{DF}^{-1}\mathbf{F}(\Bar{\mathbf{X}}^{j-1}), \ j=1,\ldots,N_{max}
\end{equation}
where $\Bar{\mathbf{X}}^j$ is the solution at iteration $j$, and $\mathbf{DF}$ is the Jacobian of the constraint vector, at iteration $j-1$.

\begin{figure}[H]
    \centering
    \includegraphics[width=.79\columnwidth]{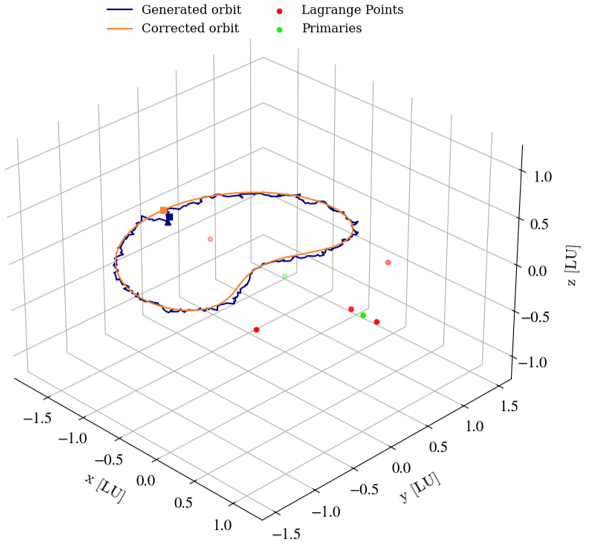}
    \caption[Example of the refinement of an orbit]{Example of a refined orbit. The blue trajectory is the initial output of the model, whereas the orange one is the trajectory refined.}
    \label{refinement_example}
\end{figure}

Convergence is achieved when the constraint vector meets an assigned tolerance. A good initial guess $\Bar{\mathbf{X}}^{0}$ is crucial for convergence. To ensure significant guesses, we set a maximum of 20 iterations ($N_{\text{max}}=20$) and used 10\% of the states equally spaced in the orbit as prompts. To visualize the effect of the refinement algorithm, we plotted an example of a generated orbit and its refinement in Figure \ref{refinement_example}.

\section{Results}

\subsection{Generation}

\begin{figure}[H]
    \centering
    \includegraphics[width=.79\columnwidth]{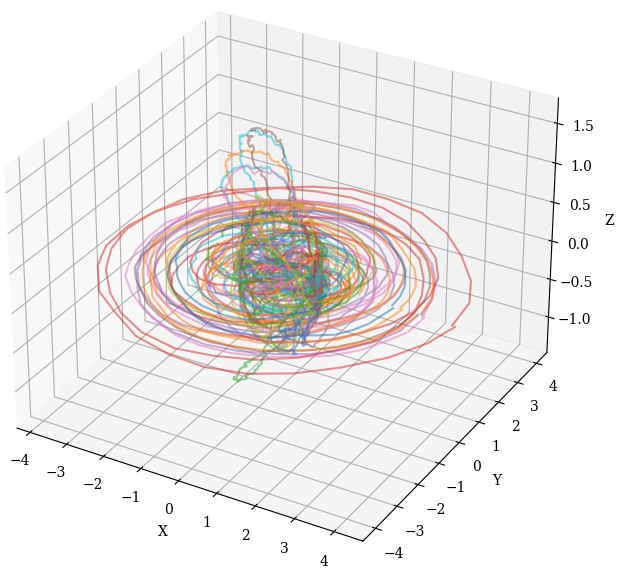}
    \caption{Generation of 100 synthetic orbits.}
    \label{fig_orbits}
\end{figure}

For this experiment, the training dataset sequence length, and consequently the timesteps of the generation, were set to 100. A VAE with 2 latent dimensions was trained on the dataset over several epochs. Although we believe that the loss may not be entirely meaningful for performance due to the refinement algorithm used at the end of the pipeline, the final model loss was 8.6, with a reconstruction loss of 2.8 and a KL loss of 5.8. After the training, 100 new orbits were generated by sampling from the latent space (Figure~\ref{fig_orbits}).

By examining the results, it is evident that the model has learned and represented orbits that closely resemble actual orbits in essence and shape. However, the generated orbits exhibit an inherent fuzziness not present in real data, resulting in the orbits appearing as fuzzy lines rather than distinct ones. This fuzziness is further corroborated by the physical error check, which shows an average error of 34 for each node, whereas a physical orbit should exhibit an error in the order of magnitude of \num{1e-13} (absolute tolerance of the ODE solver). Another important aspect is that the model consistently learned that periodic orbits need to close, meaning that the final value of the position needs to be close to the initial value.

\subsection{Latent Space}

\begin{figure}[H]
    \centering
    \includegraphics[width=0.97\columnwidth]{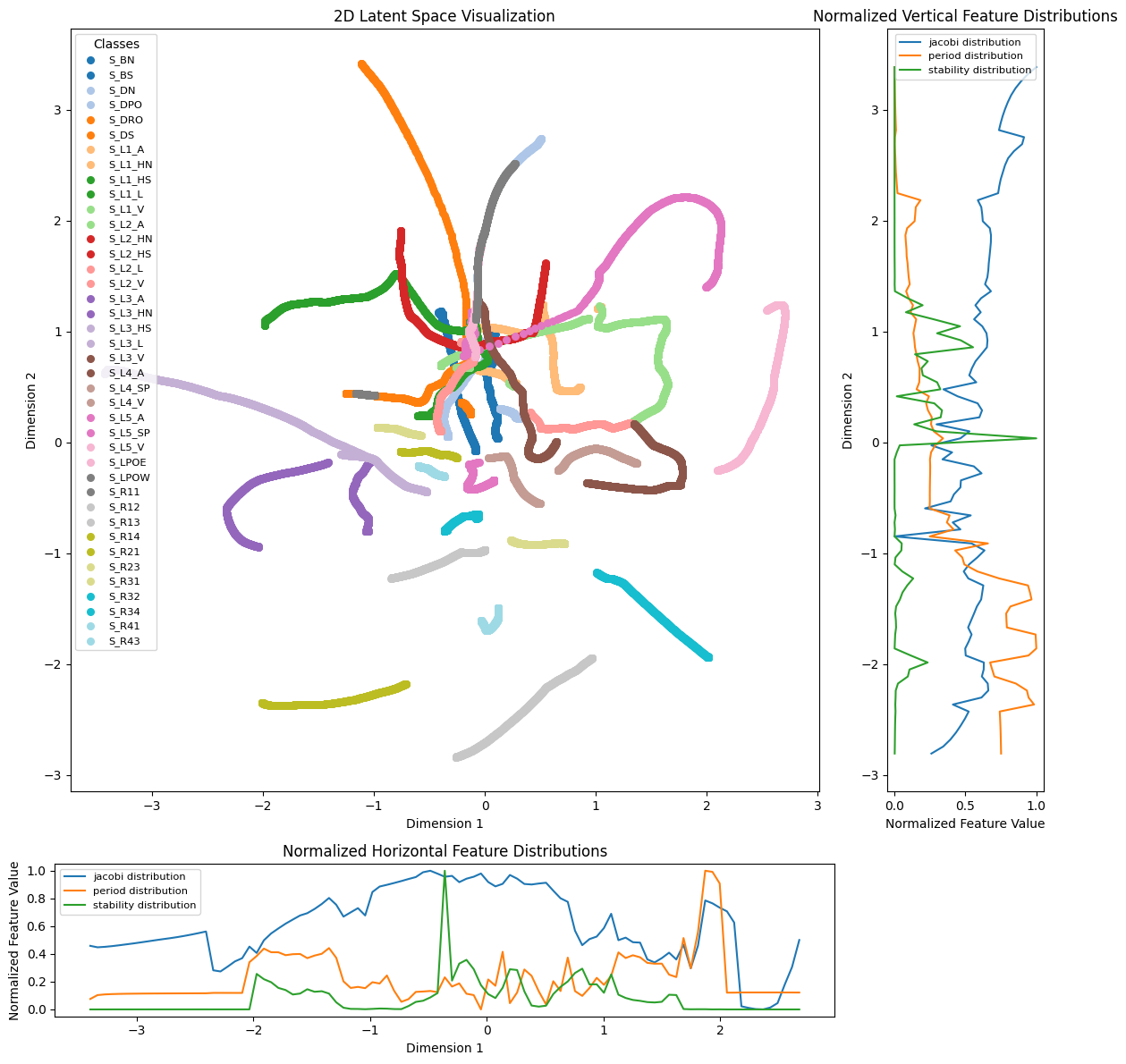}
    \caption{Visualization of the latent space from the experiment colored by orbital family labels, with plots showing the average distribution of the Jacobi constant, period, and stability.}
    \label{exp1_latent_space_distribution}
\end{figure}

We explored the latent space without the need for dimensionality reduction, as the latent dimensions were already set to two. This exploration is shown in Figure \ref{exp1_latent_space_distribution}. To properly represent each data point in the latent space, we plotted the mean of each distribution, as it is the most representative point. As expected, orbits of the same class are observed grouping together in the latent space, forming filament-like structures. This clustering occurs despite the absence of labels during training, indicating that the model has learned an internal representation of the orbit families and groups orbits with similar characteristics. To further corroborate this insight, we applied Gaussian Mixture Models to cluster the latent space and then applied certain metrics to quantify the extent of the clustering. For a clustering of 40 classes, the performance yielded a Normalized Mutual Information (NMI) score of 0.78 and an accuracy rate of 0.56.

Furthermore, we had a list of three features—Jacobi constant, period, and stability—for every orbit in the training dataset. We decided to plot the average of these features for each line in the latent space, both vertically and horizontally, effectively creating a representation of the features' distribution in the latent space (Figure \ref{exp1_latent_space_distribution}). We found that the period increases incrementally from up to down along the vertical axis. This suggests that the model has encoded the period in an unsupervised manner along this axis, reminiscent of the insight from the seminal VAE paper by Kingma and Welling \cite{kingma2013auto}, where moving along one dimension of the latent space caused an incremental change in the generated faces' orientation.

Regarding the other features, no clear correlation was found. We observed high stability in the middle of the distribution, which might indicate another type of encoding. The Jacobi constant appeared to be equally distributed throughout the entire latent space, showing no specific pattern.

\subsection{Refinement}

\begin{figure}[H]
    \centering
    \includegraphics[width=.89\columnwidth]{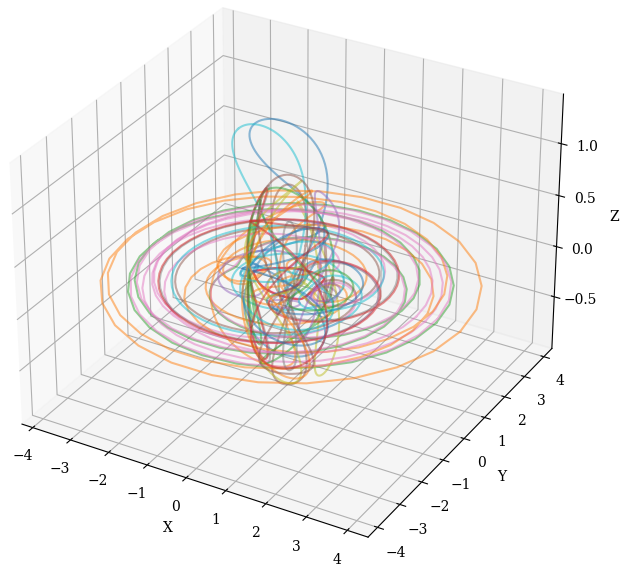}
    \caption{Successful refinement of the generated orbits.}
    \label{refined_orbits}
\end{figure}

After the generation process, the 100 synthetic orbits were refined using the convergence algorithm described in Section \ref{sec:convergence}. We found that 46 out of the 100 generated orbits were sufficiently accurate guesses for the refinement algorithm to successfully converge, achieving a ratio of convergence of 0.46 and an average of 10.1 iterations for convergence. The refined orbits, computed from the ones shown in Figure \ref{fig_orbits}, are displayed in Figure \ref{refined_orbits}. Interestingly, all of the final refined orbits were new, meaning none of them were present in the training data.

\section{Discussion}

The paper presented what can be considered a first example of Generative Astrodynamics. The results in the paper demonstrate how families of periodic orbits can be encoded into a two-dimensional latent space of features and decoded back into families of approximated periodic orbits.
By sampling the latent space we demonstrated the generation of approximated periodic trajectories that converge to physical ones after a local refinement. 
The results in this paper are evidence of the transformative potential of generative AI in astrodynamics. One particularly promising area is the use of generative models for orbit discovery. We have demonstrated the feasibility of creating a pipeline that generates physical trajectories rather than merely optimizing existing ones. This achievement represents a significant advancement not documented in the literature at the inception of this project. By exploring other AI architectures and expanding our datasets to include other planetary systems, we aim to revolutionize space mission design, minimize mission analysis costs, capture and transfer past orbital knowledge, and enable the training of specialized models through synthetic data generation.

\section{Acknowledgements}

This research was partially funded by the European Space Agency (ESA) project OrbitGPT, grant number 40001435.

\printbibliography
\addcontentsline{toc}{section}{References}

\end{document}